\documentclass[letterpaper, 10 pt, conference]{ieeeconf}  

\IEEEoverridecommandlockouts                              
\usepackage{booktabs}
\usepackage{float}
\usepackage{tensor}
\usepackage{url}
\usepackage{amsmath} 
\usepackage{amssymb}  
\usepackage{times}
\usepackage{subcaption}
\usepackage{threeparttable}
\usepackage[utf8]{inputenc}
\usepackage[overload]{empheq}
\usepackage{cases} 
\usepackage{multirow}
\usepackage{graphicx}

\title{\LARGE \bf
Efficient Trajectory Optimization for Robot Motion Planning
}

\author{Yu Zhao, Hsien-Chung Lin, and Masayoshi Tomizuka
\thanks{Yu Zhao, Hsien-Chung Lin, and Masayoshi Tomizuka are with the Department of Mechanical Engineering,
        University of California at Berkeley, Berkely, CA
94720, USA
        {\tt\small \{yzhao334,hclin\}@berkeley.edu, tomizuka@me.berkeley.edu}}%
}

\begin{document}

\maketitle
\thispagestyle{empty}
\pagestyle{empty}

\begin{abstract}

Motion planning for multi-jointed robots is challenging. Due to the inherent complexity of the problem, most existing works decompose motion planning as easier subproblems. However, because of the inconsistent performance metrics, only sub-optimal solution can be found by decomposition based approaches. This paper presents an optimal control based approach to address the path planning and trajectory planning subproblems simultaneously. Unlike similar works which either ignore robot dynamics or require long computation time, an efficient numerical method for trajectory optimization is presented in this paper for motion planning involving complicated robot dynamics. The efficiency and effectiveness of the proposed approach is shown by numerical results. Experimental results are used to show the feasibility of the presented planning algorithm.
\end{abstract}

\section{INTRODUCTION}
Motion planning for robots with multi-jointed arms is challenging. Due to the complicated geometric structure and nonlinear dynamics, time-consuming computation is required to solve motion planning problem even in the simplest cases. Most existing works utilize the path-velocity decomposition approach \cite{pham2017admissible}, in which motion planning problem is separated into easier subproblems, i.e., path planning and trajectory planning. The path planning problem focuses on the generation of collision free geometric path in the configuration space, while the trajectory planning problem focuses on the generation of time optimal velocity profile along the geometric path. Extensive research \cite{lavalle2006planning, sucan2012open, verscheure2009time, reynoso2016convex} has been conducted for each subproblem, resulting a rich collection of algorithms. However, due to the inconsistency of performance metrics between motion planning problem and the subproblems, only sub-optimal solution can be found by path-velocity decomposition based approaches. Time optimal motion planning involving collision avoidance requirement and robot dynamics is still a challenging problem (\cite{la2011motion, pham2014general}).

In order to avoid the inconsistency of performance metrics, this paper presents an optimal control based approach to address the path planning and trajectory planning problems simultaneously. The presented approach is able to generate time optimal trajectories without predetermining the geometric path while satisfying constraints involving robot dynamics. Similar works can be found in \cite{schulman2013finding, chettibi2004minimum, diehl2006fast}. However either robot dynamics are ignored or long computation time is required in these works. In this paper, an efficient numerical method for trajectory optimization is utilized to solve the optimal control problem for robot motion planning. It is shown by numerical results that the solution can be found with short computation time even when complicated robot dynamics are involved. Experimental results have shown the feasibility of the planned motion.

The rest part of this paper is organized as follows: section \ref{formulation} presents optimal control formulation for robot motion planning problems, section \ref{numerical} presents an efficient numerical method for trajectory optimization, section \ref{example} presents numerical and experimental results of the proposed approach, and section \ref{conclusion} concludes this paper.

\section{Problem Formulation}\label{formulation}
A general optimal control problem can be posed as follows: determine the state-control function pair, $t\mapsto(\pmb{x},\pmb{u})$, terminal time $t_f$, that minimize the performance metric or \textit{cost function}, while satisfying \textit{dynamic constraints}, \textit{path constraints}, and \textit{boundary conditions}  (\cite{ross2012review}). The robot motion planning problem can be formulated as an optimal control problem by defining the cost function, dynamic constraints, path constraints, and boundary conditions. 

The state and control in motion planning involving robot dynamics can be defined as:
\begin{equation}
\pmb{x}(t)=\begin{bmatrix}
\pmb{q}(t)\\
\dot{\pmb{q}}(t)
\end{bmatrix},\quad
\pmb{u}(t)=\pmb{\tau}(t)
\end{equation}
where $t\in[0,t_f]$, $\pmb{q}(t)=\left[q_1(t),\cdots,q_n(t)\right]^T$ is the vector for joint positions, and $\pmb{\tau}(t)=\left[\tau_1(t),\cdots,\tau_n(t)\right]^T$ is the vector for joint torques, $n$ is the number of robot joints.

\subsection{Cost Function}
The quality of the planned motion strongly depends on the formulation of cost function. In this paper, the cost function is formulated as a summation of motion time $t_f$ and a regularization term for smoothness and naturalness of the generated motion:
\begin{equation}
J=t_f+\mu \int_{0}^{t_f}\dddot{\pmb{q}}(t)^T\pmb{Q}\dddot{\pmb{q}}(t)\mathrm{d}t
\end{equation}
where $\dddot{\pmb{q}}(t)$ is the jerk of joint motion. The regularization term is designed based on the minimum-jerk model of human motion (\cite{flash1985coordination}) and thus corresponds to the importance of the naturalness of the generated motion. $\mu\geq0$ is a weighting coefficient for the regulation term, and $\pmb{Q}$ is a weight matrix designed to penalize the motion of joints with higher gear ratios. The weight matrix $\pmb{Q}$ is defined as 
\begin{equation}
\pmb{Q}(i,j)=\begin{cases}
0,& j\neq i\\
1\left/ \pmb{R}(i,i)^2\right. ,& j=i
\end{cases}
\end{equation}

The case $\mu=0$ corresponds to the time optimal motion planning problem. Larger $\mu$ slows down the generated motion, but increases the naturalness and smoothness.

\subsection{Dynamic Constraints}
The dynamic constraints is the robot dynamics. The equations of motion can be derived using Lagrangian's equations or Newton-Euler approach:
\begin{equation}
\pmb{M}(\pmb{q})\ddot{\pmb{q}}+\pmb{C}(\pmb{q},\dot{\pmb{q}})\dot{\pmb{q}}+\pmb{G}(\pmb{q})+\pmb{f}^{f}=\pmb{\tau}
\end{equation}
where $\pmb{M}(\pmb{q})\in\mathbb{R}^{n\times n}$ is the inertia matrix, $\pmb{C}(\pmb{q},\dot{\pmb{q}})\dot{\pmb{q}}\in\mathbb{R}^{n\times 1}$ is the Coriolis and centrifugal force term, $\pmb{G}(\pmb{q})\in\mathbb{R}^{n\times 1}$ is the gravity term, and $\pmb{f}^{f}\in\mathbb{R}^{n\times 1}$ is the friction term. For multi-jointed robot arms, all of these terms are inherently nonlinear.

Letting $\pmb{x}_1=\pmb{q}$, $\pmb{x}_2=\dot{\pmb{q}}$, the dynamic constraints can be formulated as state space model with state $\pmb{x}=[\pmb{x}_1^T,\pmb{x}_2^T]^T$ and control $\pmb{u}=\pmb{\tau}$:
\begin{subequations}
\begin{align}[left = \empheqlbrace\,]
&\dot{\pmb{x}}_1=\pmb{x}_2\\
&\dot{\pmb{x}}_2=\pmb{M}(\pmb{x}_1)^{-1}\left[
\pmb{u}-\pmb{C}(\pmb{x}_1,\pmb{x}_2)\pmb{x}_2-\pmb{G}(\pmb{x}_1)-\pmb{f}^{f}
\right]\label{dynEqnAcc}
\end{align}
\end{subequations}

The state space model can be rewritten as:
\begin{equation}\label{statespace}
\frac{\mathrm{d}}{\mathrm{d}t}\pmb{x}=\pmb{F}(\pmb{x},\pmb{u})
\end{equation}

\subsection{Path Constraints}
A set of path constraints can be formulated to accommodate various physical limitations of the robot actuators, as well as collision free conditions. The path constraints for robot motion planning include:
\begin{subequations}\label{6jntPathconstraint}
\begin{align}
&\textrm{Position bounds:} && \pmb{q}_{min}\leq \pmb{q}(t)\leq \pmb{q}_{max}\\
&\textrm{Velocity bounds:} && \dot{\pmb{q}}_{min}\leq \dot{\pmb{q}}(t)\leq \dot{\pmb{q}}_{max}\\
&\textrm{Torque bounds:} && \pmb{\tau}_{\min}\leq 
\pmb{\tau}(t)
\leq \pmb{\tau}_{\max}\\
&\textrm{Torque rate bounds:} && \dot{\pmb{\tau}}_{\min}\leq
\dot{\pmb{\tau}}(t)
\leq \dot{\pmb{\tau}}_{\max}
\end{align}
\end{subequations}

\begin{figure}[ht]
\begin{center}
\includegraphics[height=.269\textwidth]{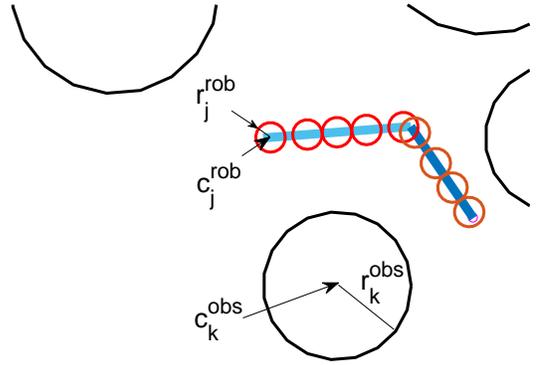}
\caption{Sphere approximation of robot and obstacles}
\label{sphereTree}
\end{center}
\end{figure}

Collision free conditions are also included in the path constraints. Dues to the complicated geometric mapping between robot workspace and configuration space, it is difficult to represent collision free conditions analytically. To simplify the formulation, the robot links and obstacles can be approximated by a set of spheres for differentiable collision detection, as illustrated in Fig. \ref{sphereTree}. The approximation can be performed either manually or automatically using sphere-tree construction algorithms \cite{bradshaw2004adaptive}.
Suppose $M$ spheres are used to approximate robot links, and $S$ spheres are used to approximate obstacles. Let the center and radius of each sphere that representing robot links be $\pmb{c}_j^{rob}(\pmb{q}(t)), r_j^{rob}, j=1,\cdots,M$, the center and radius of each sphere that representing obstacles be $\pmb{c}_k^{obs}, r_k^{obs}, k=1,\cdots,S$. Robot forward kinematics problem can be solved to determine the functional relationship between joint positions $\pmb{q}(t)$ and the location of sphere centers $\pmb{c}_j^{rob}(\pmb{q}(t))$. The collision free constraints can then be formulated as
\begin{subequations}\label{collision}
\begin{align}
&\textrm{Self:} && 
\|\pmb{c}_j^{rob}-\pmb{c}_k^{rob}\|_2 \geq r_j^{rob}+r_k^{rob},\, [j,k]\in I\\
&\textrm{Obstacle:} && 
\|\pmb{c}_j^{rob}-\pmb{c}_l^{obs}\|_2 \geq r_j+r_l^{obs},\,\forall j,l
\end{align}
\end{subequations}
where $I$ is a set of indices indicating possible collision between two balls that approximate robot links.

When workspace boundaries are presented, additional path constraints are necessary. Let $[x_{\min}^{bnd},x_{\max}^{bnd}]$, $[y_{\min}^{bnd},y_{\max}^{bnd}]$, and $[z_{\min}^{bnd},z_{\max}^{bnd}]$ be the workspace limits for $X$, $Y$, and $Z$ directions respectively. Let $\pmb{c}_{\min}=[x_{\min}^{bnd},y_{\min}^{bnd},z_{\min}^{bnd}]^T$, $\pmb{c}_{\max}=[x_{\max}^{bnd},y_{\max}^{bnd},z_{\max}^{bnd}]^T$. The path constraints for workspace boundary can be formulated as:
\begin{subequations}\label{workspace}
\begin{align}
&\textrm{Workspace lower bound:} && \pmb{c}_{\min} \leq \pmb{c}_j^{rob}-r_j^{rob},\,\forall j\\
&\textrm{Workspace upper bound:} && 
\pmb{c}_{\max} \geq \pmb{c}_j^{rob}+r_j^{rob},\,\forall j
\end{align}
\end{subequations}

\subsection{Boundary Conditions}

The boundary conditions for robot motion planning problem include:
\begin{subequations}
\begin{align}
&\textrm{Initial \& final position} && \pmb{q}(0) = \pmb{q}^0, \pmb{q}(t_f) = \pmb{q}^f\\
&\textrm{Initial \& final velocity} && \dot{\pmb{q}}(0)=0,  \dot{\pmb{q}}(t_f)=0\\
&\textrm{Initial \& final acceleration} && \ddot{\pmb{q}}(0)=0,  \ddot{\pmb{q}}(t_f)=0\\
&\textrm{Terminal time bounds :} && t_f^{\min}\leq t_f\leq t_f^{\max}
\end{align}
\end{subequations}
where $\pmb{q}^0$ and $\pmb{q}^f$ are the initial and target joint positions, $t_f^{\min}$ and $t_f^{\max}$ are the minimum and maximum allowed terminal time.

\section{Efficient Numerical Method for Trajectory Optimization}\label{numerical}
Trajectory optimization is a technique for computing an open-loop solution to an optimal control problem. Since no universal analytical solution can be found for nonlinear optimal control problems, a variety of numerical approaches have been developed for trajectory optimization in \cite{rao2009survey,betts2010practical}. In most numerical approaches, the continuous time optimal control problem is firstly converted into discretized optimization problem in a procedure called transcription (\cite{kelly2015transcription}). The optimization problem is then solved by general purpose optimization solver. Polynomial interpolation is finally utilized to return an approximate solution to the continuous time optimal control problem.

\begin{figure*}[!ht]
\centering
\includegraphics[height=.35\textwidth]{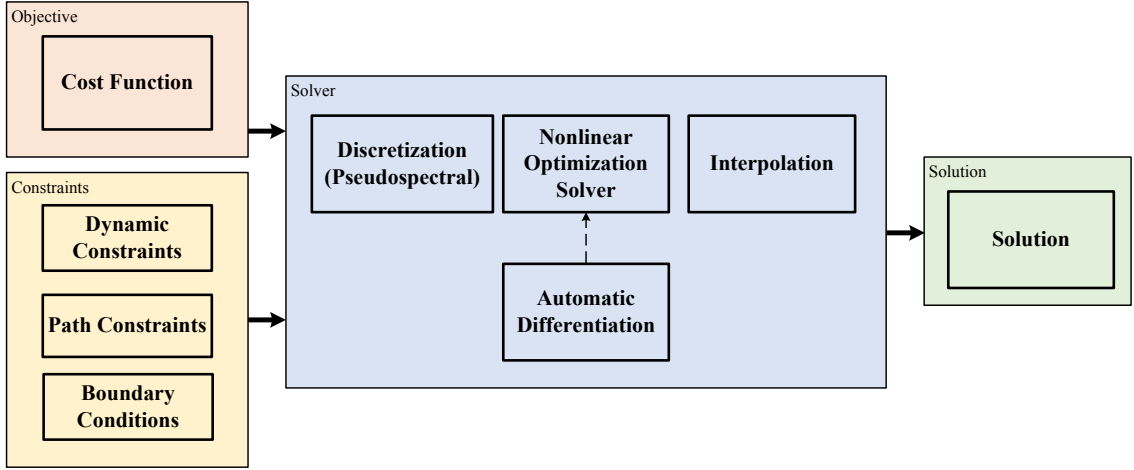}
\caption{Efficient numerical method for trajectory optimization}
\label{framework}
\end{figure*}

In numerical methods for trajectory optimization, two parts are playing the key role: discretization and optimization. An efficient implementation can be designed by choosing these two components intelligently. In this paper, the pseudospectral method is chosen to transcribe the continuous time optimal control problem, and the interior point method with the support of automatic differentiation is chosen to solve the discretized optimization problem, as illustrated in Fig. \ref{framework}.

\subsection{Pseudospectral Method}
The decision variables in the continuous time optimal control problem for robot motion planning include $t_f$, $\pmb{x}(t)$, and $\pmb{u}(t)$. In the transcription procedure, $\pmb{x}(t)$ and $\pmb{u}(t)$ are discretized by their values at certain time as $\{\pmb{x}(T_i),i=0,\cdots N\}$ and $\{\pmb{u}(T_i),i=0,\cdots N\}$, where $\{T_i,i=0,\cdots N\}$ are called knots, and $N$ is the number of knots. It is reported in previous research \cite{rao2009survey, kelly2017introduction, trefethen2013approximation} that the solution accuracy increases exponentially fast with the increase of interpolation knots for pseudospectral methods. Thus high computational efficiency can be achieved using pseudospectral method since less discretization knots can be chosen under the same solution accuracy requirement. In addition, the approximate solution is guaranteed to be smooth since high order global polynomial interpolation is utilized in pseudospectral methods.

In this paper, the Chebyshev-Lobatto points (or Chebyshev points) are chosen to be the knots. Such choice can avoid the oscillation phenomenon in high order global polynomial interpolation. For $t\in[0,t_f]$, the knots are:
\begin{equation}
T_i=\frac{t_f}{2}\left[\cos\left(\frac{i\pi}{N}\right)+1\right],i=0,\cdots,N
\end{equation}

Pseudospectral methods have provided a set of tools for polynomial interpolation, approximating integration terms using quadrature, and approximating derivatives using differential matrix.

\begin{enumerate}
\item{
\textbf{Interpolation}
The polynomial interpolation in pseudospectral methods can be performed by \textit{barycentric interpolation}, which can be formulated as a linear combination of Lagrangian polynomials. For state trajectory and control trajectory, the form is
\begin{equation}
\begin{array}{l}
\displaystyle \pmb{x}(t)\approx\sum\limits_{j=0}^{N}\pmb{x}(T_j)\ell_j(t)\\
\displaystyle \pmb{u}(t)\approx\sum\limits_{j=0}^{N}\pmb{u}(T_j)\ell_j(t)
\end{array}
\label{interpolation}
\end{equation}
where $\ell_j(t)$ is the $j$th Lagrange polynomial. In barycentric interpolation, a special form of Lagrange polynomial is implemented to efficiently perform interpolation (\cite{trefethen2013approximation, berrut2004barycentric}).

}
\item{
\textbf{Quadrature}
Quadrature is the standard term for the numerical calculation of integrals. The integration of function $\mathcal{L}[\pmb{x}]$ can be approximately evaluated by quadrature rules in pseudospectral methods as:
\begin{equation}
\begin{array}{rl}
\displaystyle\int_{0}^{t_f}\mathcal{L}[\pmb{x}]\mathrm{d}t
&\approx
\displaystyle\int_{0}^{t_f}\left[\sum\limits_{j=0}^{N}\ell_j(t)\mathcal{L}[\pmb{x}(T_j)]\right]\mathrm{d}t\\
&=
\displaystyle\sum\limits_{j=0}^{N}
w_j
\mathcal{L}[\pmb{x}(T_j)]
\end{array}
\end{equation}
where $\{w_j,j=0,\cdots,N\}$ is a set of quadrature weights. The quadrature weights can be explicitly defined to be
\begin{equation}
w_j=\int_{0}^{t_f}\ell_j(t)\mathrm{d}t, j=0,\cdots,N
\label{quadratureW}
\end{equation}

When Chebyshev points are chosen, the corresponding quadrature rule (Clenshaw–Curtis quadrature) can be found in \cite{waldvogel2006fast}:
}
\item{
\textbf{Differentiation matrix}
Let the stacked state and robot dynamics at knots be
\begin{equation}
\pmb{X}=\begin{bmatrix}
\pmb{x}(T_0)^T\\
\vdots\\
\pmb{x}(T_N)^T
\end{bmatrix},
\pmb{\mathcal{F}}=\begin{bmatrix}
\pmb{F}(\pmb{x}(T_0),\pmb{u}(T_0))^T\\
\vdots\\
\pmb{F}(\pmb{x}(T_N),\pmb{u}(T_N))^T
\end{bmatrix}
\end{equation}

The dynamic constraints can be posed as \cite{rao2009survey}:
\begin{equation}
\pmb{D}\pmb{X}=\frac{t_f}{2}\pmb{\mathcal{F}}
\end{equation}
where $\pmb{D}$ is the differential matrix that is used to compute the scaled time derivative of the polynomial approximation of $\pmb{x}$ (\cite{berrut2004barycentric}). Let the stacked joint torques be $\pmb{U}=[\pmb{u}(T_0),\cdots,\pmb{u}(T_N)]^T$, the stacked torque rates $\pmb{U}_d=[\dot{\pmb{u}}(T_0),\cdots,\dot{\pmb{u}}(T_N)]^T$ can be approximately calculated as:
\begin{equation}
\pmb{U}_d\approx\frac{2}{t_f}\pmb{D}\pmb{U}
\end{equation}

}
\end{enumerate}

\subsection{Automatic Differentiation}
Lots of optimization solvers are based on gradient descent algorithm. Derivative of objective function and constraints are frequently evaluated by numerical differentiation approaches, which perturbs input to the function in each dimension to obtain an approximation of the derivative using finite differences. However, numerical differentiation approaches are computationally expensive for functions with high dimensional input, and inevitably introduces round-off errors. Symbolic differentiation is one way to avoid round-off errors, however it frequently leads to inefficient code. Both numerical differentiation and symbolic differentiation are problematic in the calculation of higher order derivatives like Hessian.

To address the problems in numerical differentiation and symbolic differentiation, automatic differentiation is introduced. Automatic differentiation is a set of techniques to evaluate derivative of a function (\cite{griewank2008evaluating}). The computational cost of automatic differentiation is lower than numerical differentiation or symbolic differentiation. A rich collection of automatic differentiation implementations can be found in \cite{walther2009getting, bell2012cppad, bendtsen1996fadbad}. In this paper, CasADi \cite{Andersson2018} is chosen for its good usability in MATLAB environment. Since computational cost of automatic differentiation is proportional to that for function evaluation, articulated body algorithm \cite{featherstone2014rigid} is utilized in this work for efficient evaluation of robot dynamics.

\section{Numerical and Experimental Results}\label{example}
\begin{figure}[ht]
\centering
\includegraphics[width=.45\textwidth]{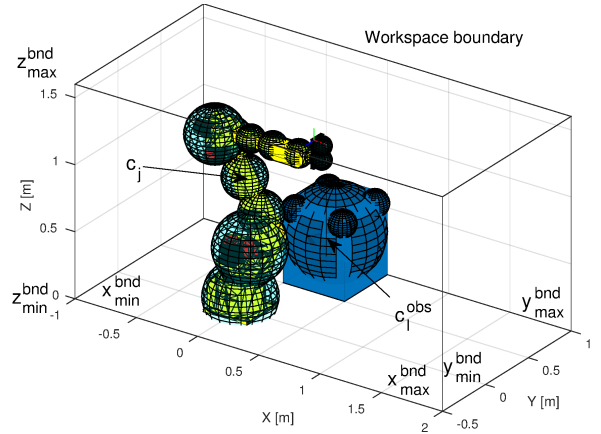}
\caption{Sphere approximation of 6-axis robot and obstacle}
\label{6jntsphere}
\end{figure}

Motion planning of a 6-axis industrial robot with dynamic constraints is considered as an example. 
\begin{figure}[!ht]
\centering
\includegraphics[width=.45\textwidth]{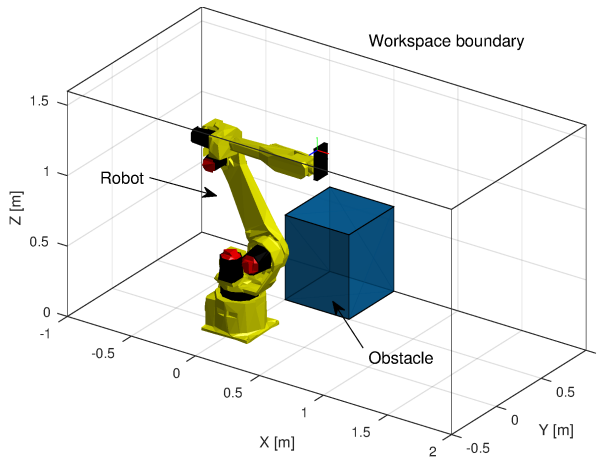}
\caption{Geometric model of a 6-axis industrial robot}
\label{6jnt}
\end{figure}
The industrial robot is supposed to work in a constrained workspace for pick-and-place tasks. The geometric model of the industrial robot is shown in Fig. \ref{6jnt}. The path constraints include joint position, velocity, torque, torque rate bounds, and collision free conditions. The sphere approximation of robot links, obstacles, and workspace boundary is shown in Fig. \ref{6jntsphere}. The actuator limits are listed in Table \ref{6jntlimits}. The actuator limitations are designed to be conservative for safety.

\begin{table}[ht]
  \centering
  \caption{Actuator limits of 6-axis industrial robot}
  \label{6jntlimits}
  \begin{tabular}{ l c c c c c c}
    \toprule
     Limits& J1 & J2 & J3 & J4 & J5 & J6    \\  \midrule
    $\pmb{q}_{\min}[^\circ]$ &
    -160 & -90 & -120 & -180 & -120 & -180\\
	$\pmb{q}_{\max}[^\circ]$ &
    170 & 90 & 230 & 180 & 100 & 180 \\
    $\dot{\pmb{q}}[^\circ/s]$ &
    $\pm$165 & $\pm$165 & $\pm$175 & $\pm$350 & $\pm$340 & $\pm$520 \\ 
	$\pmb{\tau}$[Nm] &
    $\pm$1397 & $\pm$1402 & $\pm$383 & $\pm$45.2 & $\pm$44.6 & $\pm$32.5\\
    $\dot{\pmb{\tau}}$[Nm/s] & 
    $\pm$20948 & $\pm$21035 & $\pm$5741 & $\pm$678 & $\pm$669 & $\pm$488 \\ 
    \bottomrule
  \end{tabular}
\end{table}

\subsection{Numerical Results}
Several tests have been performed to evaluate the effect of different regularization weights and number of knots. It is observed that the geometric path can be adjusted automatically for collision avoidance, even if infeasible initialization is provided. Increasing $\mu$ is helpful for shortening the computation time, but results in slower motion. Shorter motion time $t_f$ can be obtained by using more knots, but the computation takes longer time. When 12 knots are chosen, the planned motion is reasonably close to the possible optimal solution with an acceptable computation time. Less knots can be used if shorter computation time is required. $\mu=0.3$ can be chosen to balance the time-optimality requirement and the naturalness of the generated motion.
\begin{figure}[!ht]
\centering
\begin{subfigure}[b]{.35\textwidth}
		\centering
        \includegraphics[width=1\textwidth]{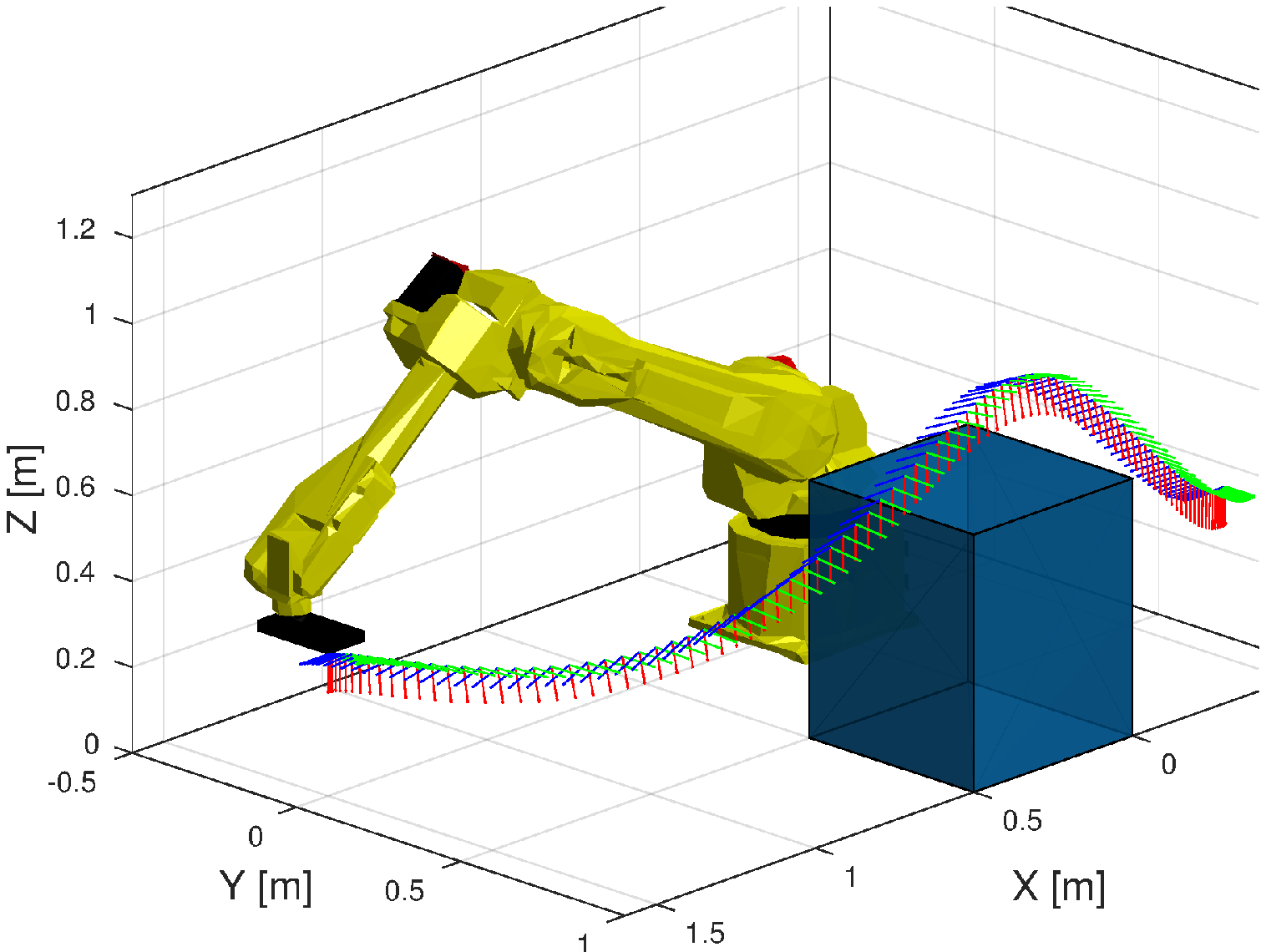}
        \subcaption{Test case 1}        
\end{subfigure}
\begin{subfigure}[b]{.35\textwidth}
		\centering
        \includegraphics[width=1\textwidth]{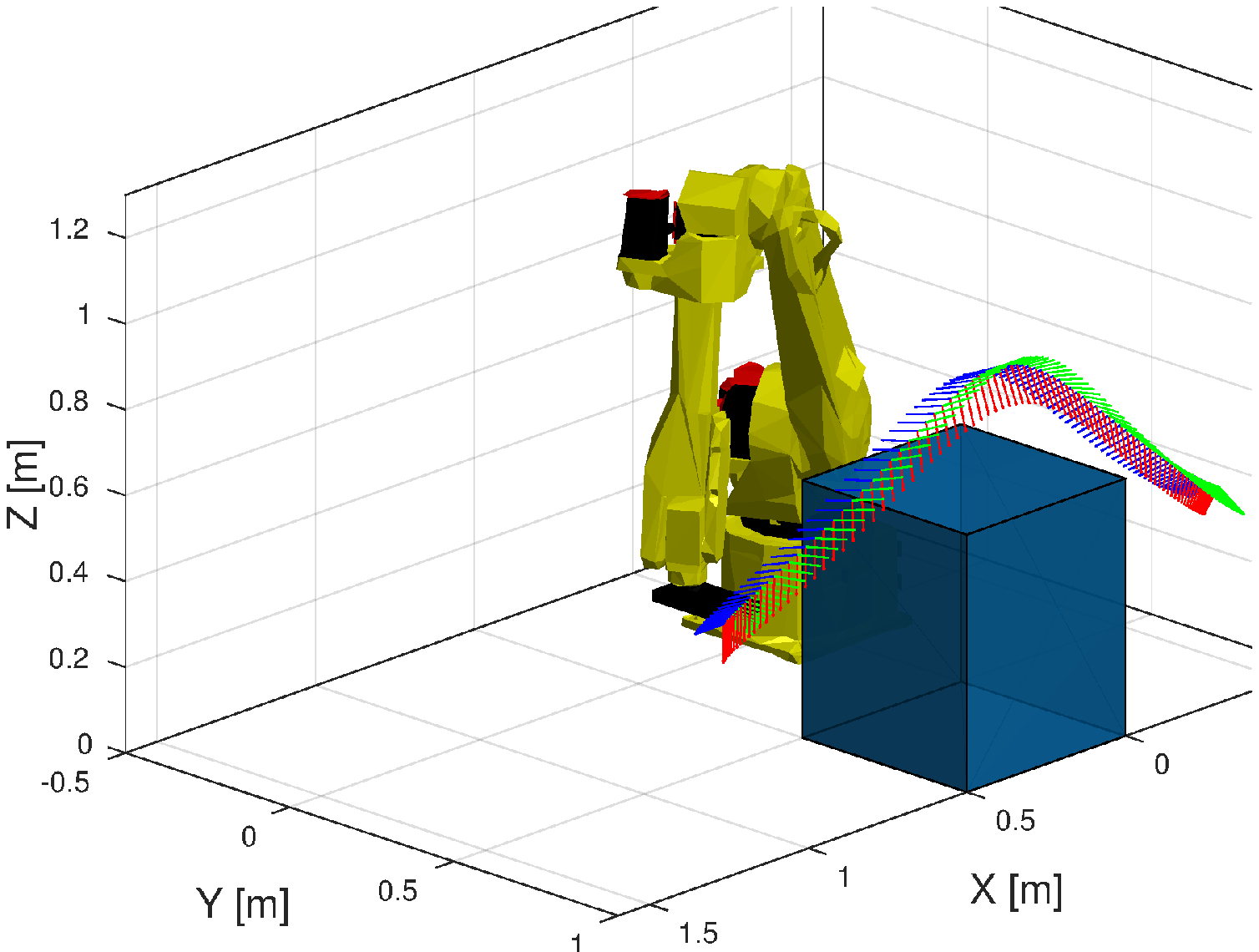}
        \subcaption{Test case 2}
\end{subfigure}
\begin{subfigure}[b]{.35\textwidth}
		\centering
        \includegraphics[width=1\textwidth]{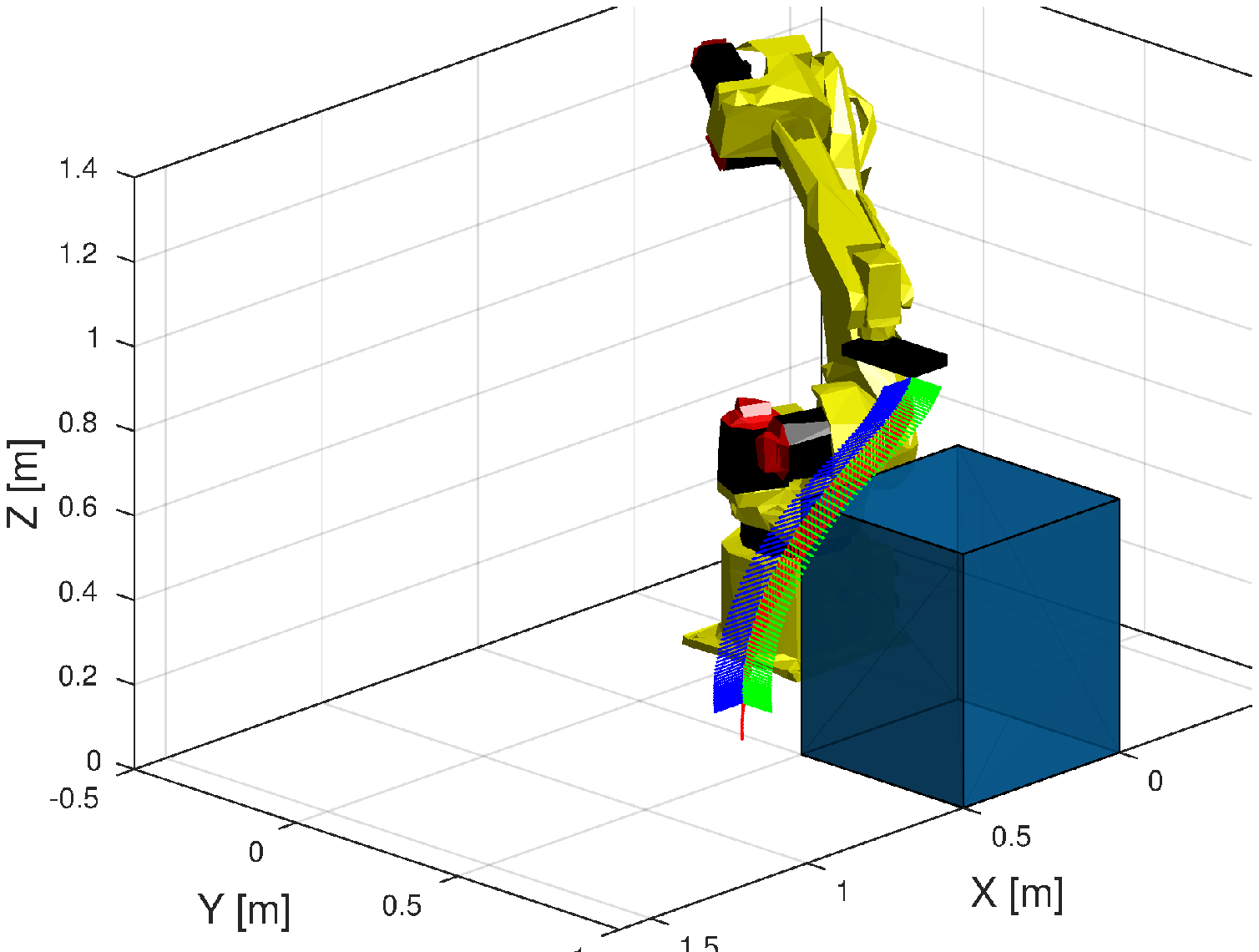}
        \subcaption{Test case 3}        
\end{subfigure}
\begin{subfigure}[b]{.35\textwidth}
		\centering
        \includegraphics[width=1\textwidth]{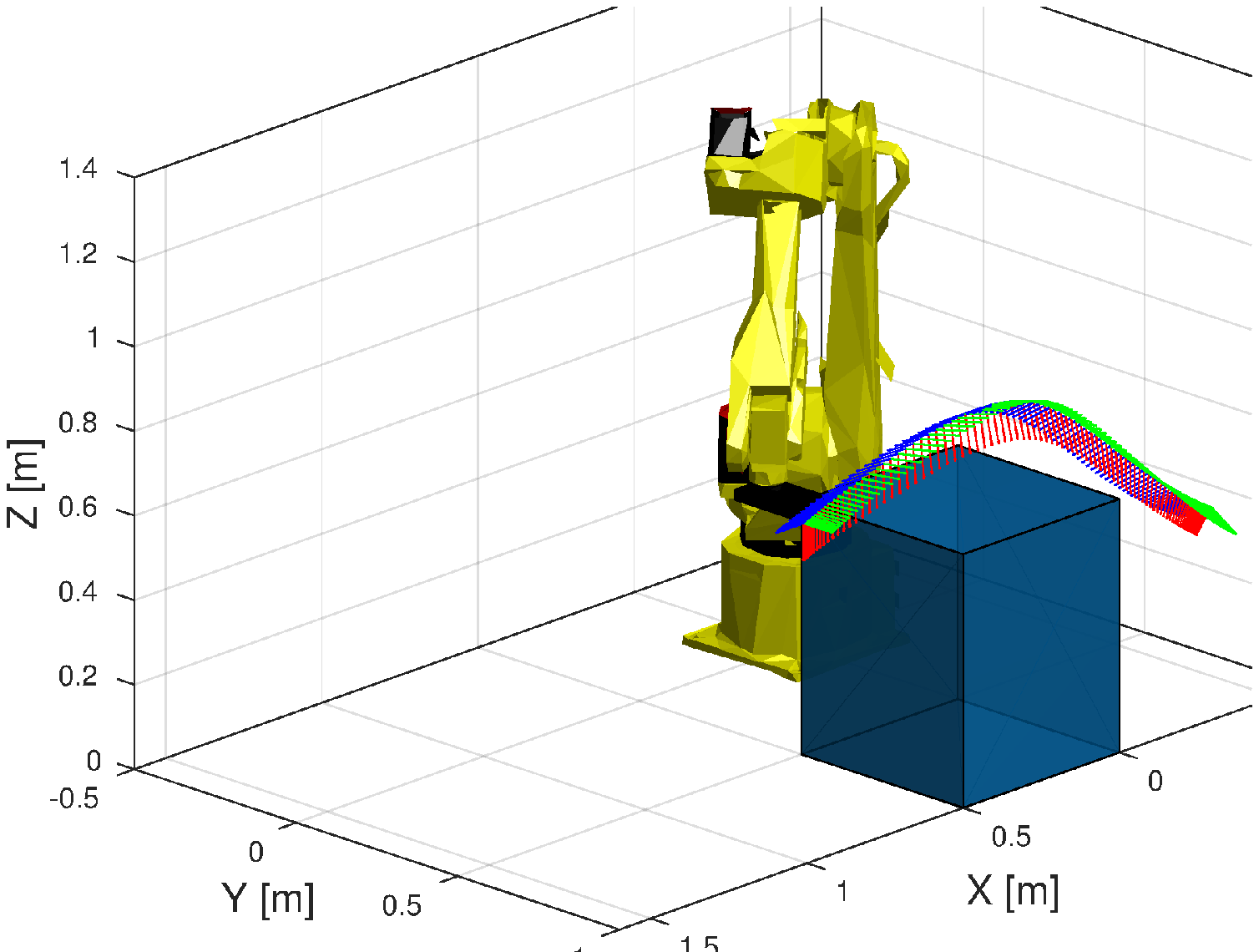}
        \subcaption{Test case 4}
\end{subfigure}
\caption{Optimal multiple joint robot trajectory with different initial and target positions, 12 knots, $\mu=0.3$}
\label{6jnttestInitTarget}
\end{figure}

The proposed planning algorithm has also been tested using a group of different initial and target positions, as shown in Fig. \ref{6jnttestInitTarget}. All the computation is performed on a laptop with a 2.1 GHz Intel\textregistered{} Core\texttrademark{} processor. The solver takes around 5.9 s for a one-time initialization for automatic differentiation by CasADi. The computation times and optimized motion times for the test cases are summarized in Table \ref{timesum}. 
\begin{table}[ht]
  \begin{center}
  \caption{Motion time and computation time for test cases}
  \label{timesum}
  \begin{tabular}{ c | c | c | c | c | c }
    \hline
    \hline
    time [s] & knots & case 1 & case 2 & case 3 & case 4    \\
    \hline
    \multirow{2}{*}{motion time $t_f$ } &
    12 &
    1.65 & 1.23 & 1.05 & 1.08 \\
    \cline{2-6}
    & 8 &
    1.98 & 1.23 & 1.09 & 1.11 \\
    \hline
    \multirow{2}{*}{computation time } &
    12 &
    2.55 & 2.37 & 2.39 & 2.57 \\
    \cline{2-6}
    & 8 &
    1.28 & 1.75 & 0.84 & 1.32 \\
    \hline
  \end{tabular}
  \end{center}
\end{table}
In all the test cases, good approximations of the time optimal trajectories are returned in about 2-3 seconds using 12 knots, and about 1-2 seconds using 8 knots. Existing works \cite{chettibi2004minimum, diehl2006fast} require from 20 seconds to several minutes for computation, in which only robot dynamics are involved but not collision avoidance. The proposed approach is highly efficient comparing to these results.

\subsection{Experimental Results}
\begin{figure}[ht]
\centering
\begin{subfigure}[b]{.22\textwidth}
		\centering
        \includegraphics[width=1\textwidth]{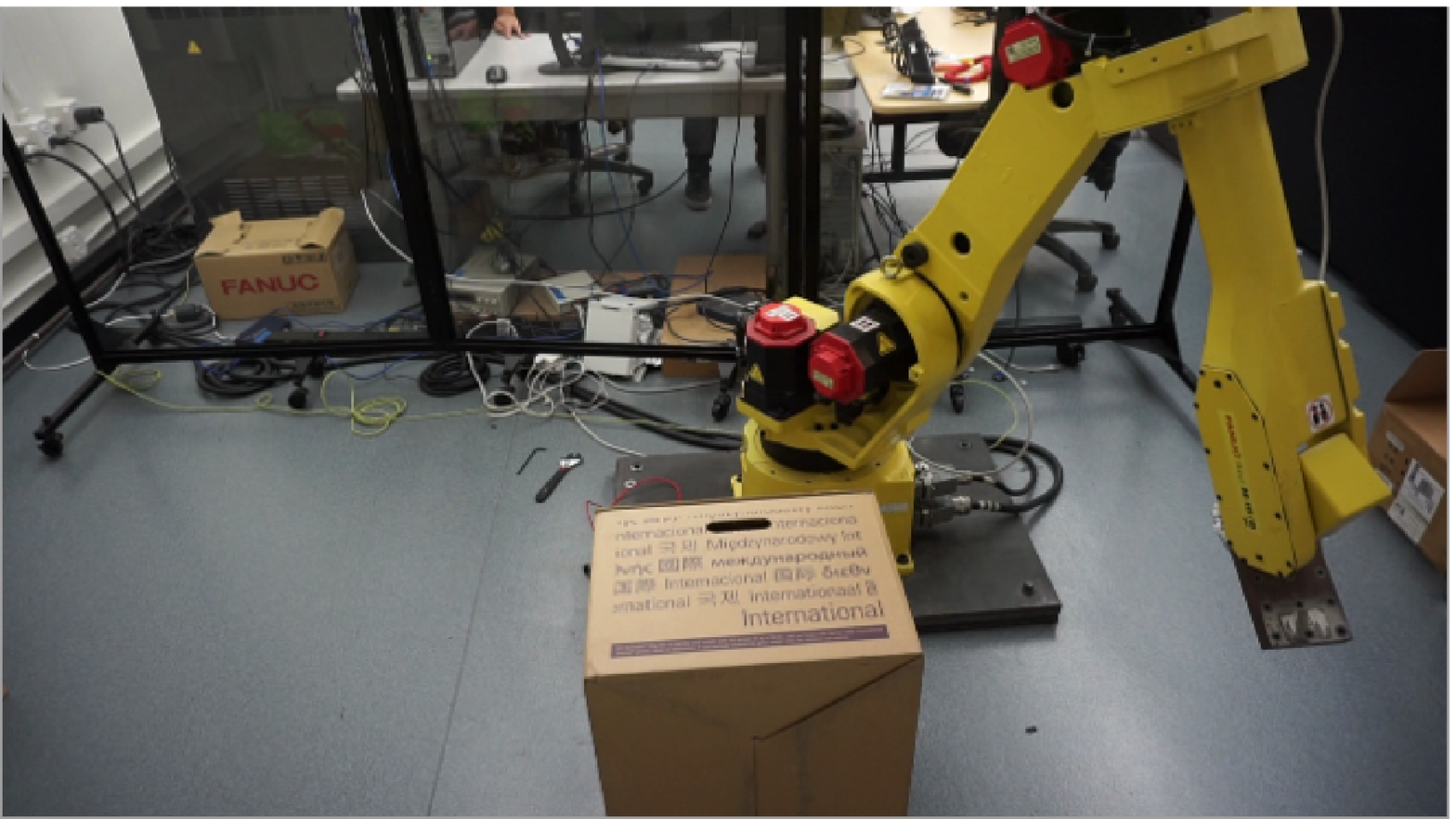}
        \subcaption{t = 0 s}        
\end{subfigure}
\begin{subfigure}[b]{.22\textwidth}
		\centering
        \includegraphics[width=1\textwidth]{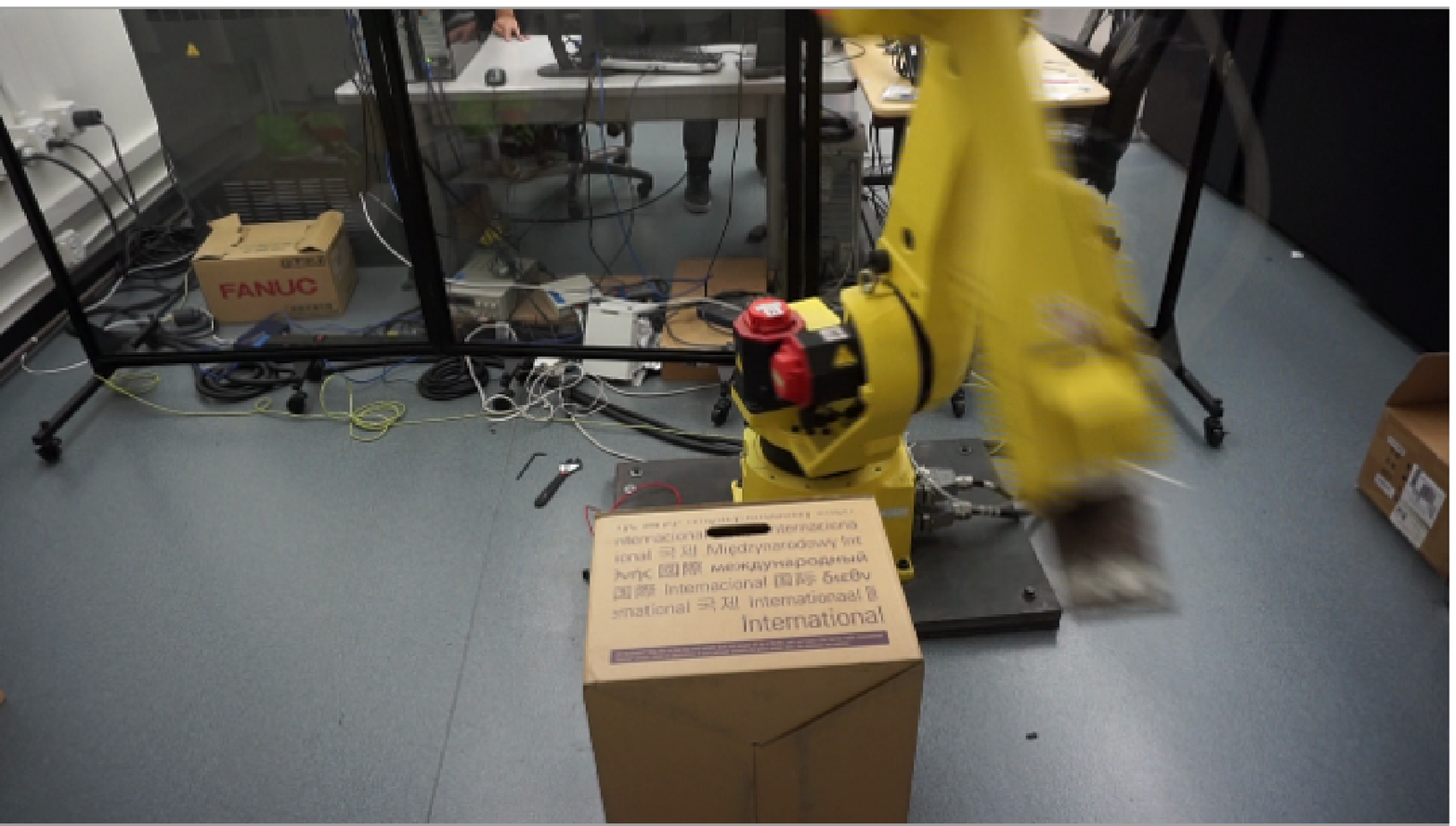}
        \subcaption{t = 0.33 s}
\end{subfigure}
\begin{subfigure}[b]{.22\textwidth}
		\centering
        \includegraphics[width=1\textwidth]{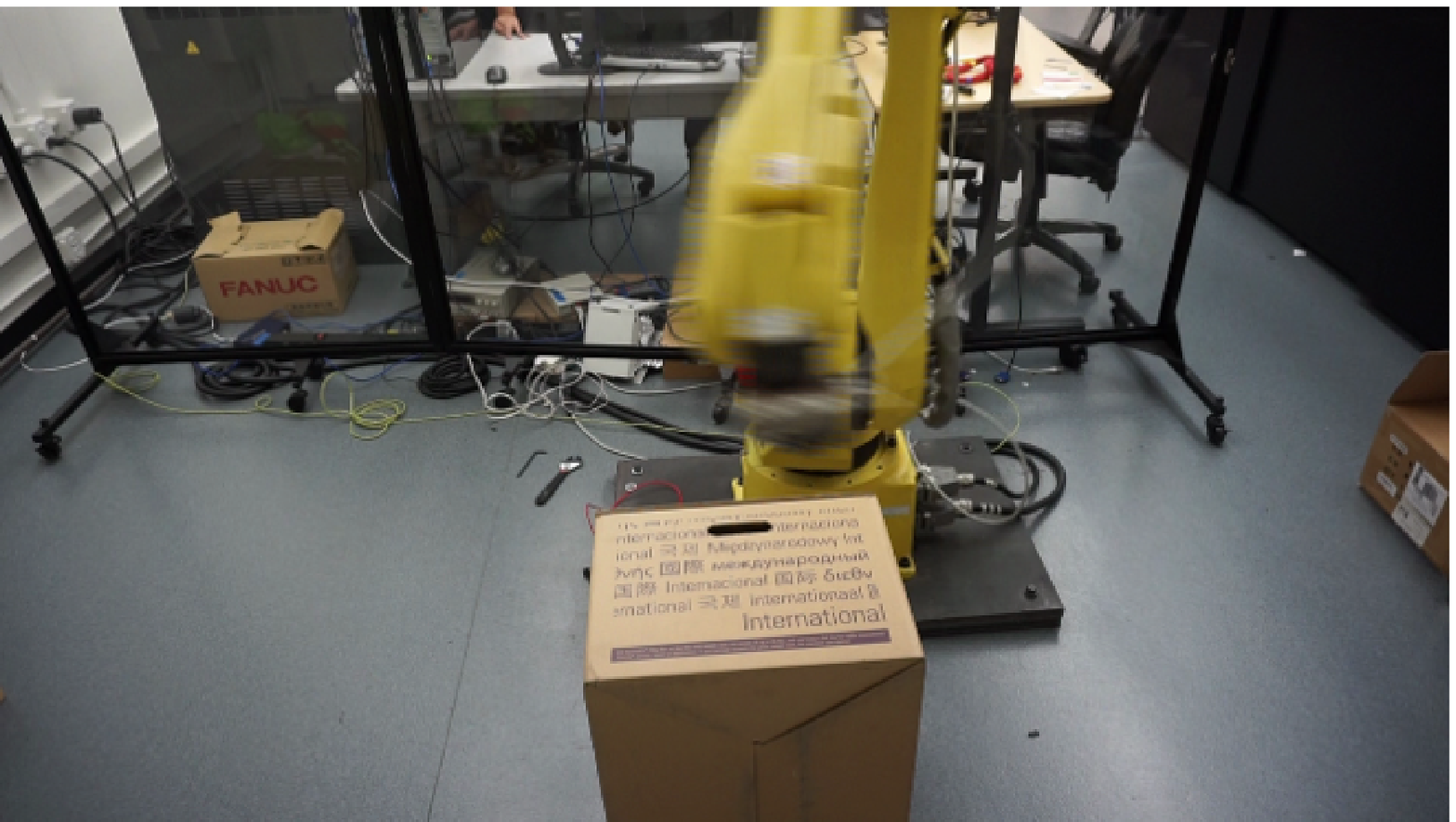}
        \subcaption{t = 0.56 s}
\end{subfigure}
\begin{subfigure}[b]{.22\textwidth}
		\centering
        \includegraphics[width=1\textwidth]{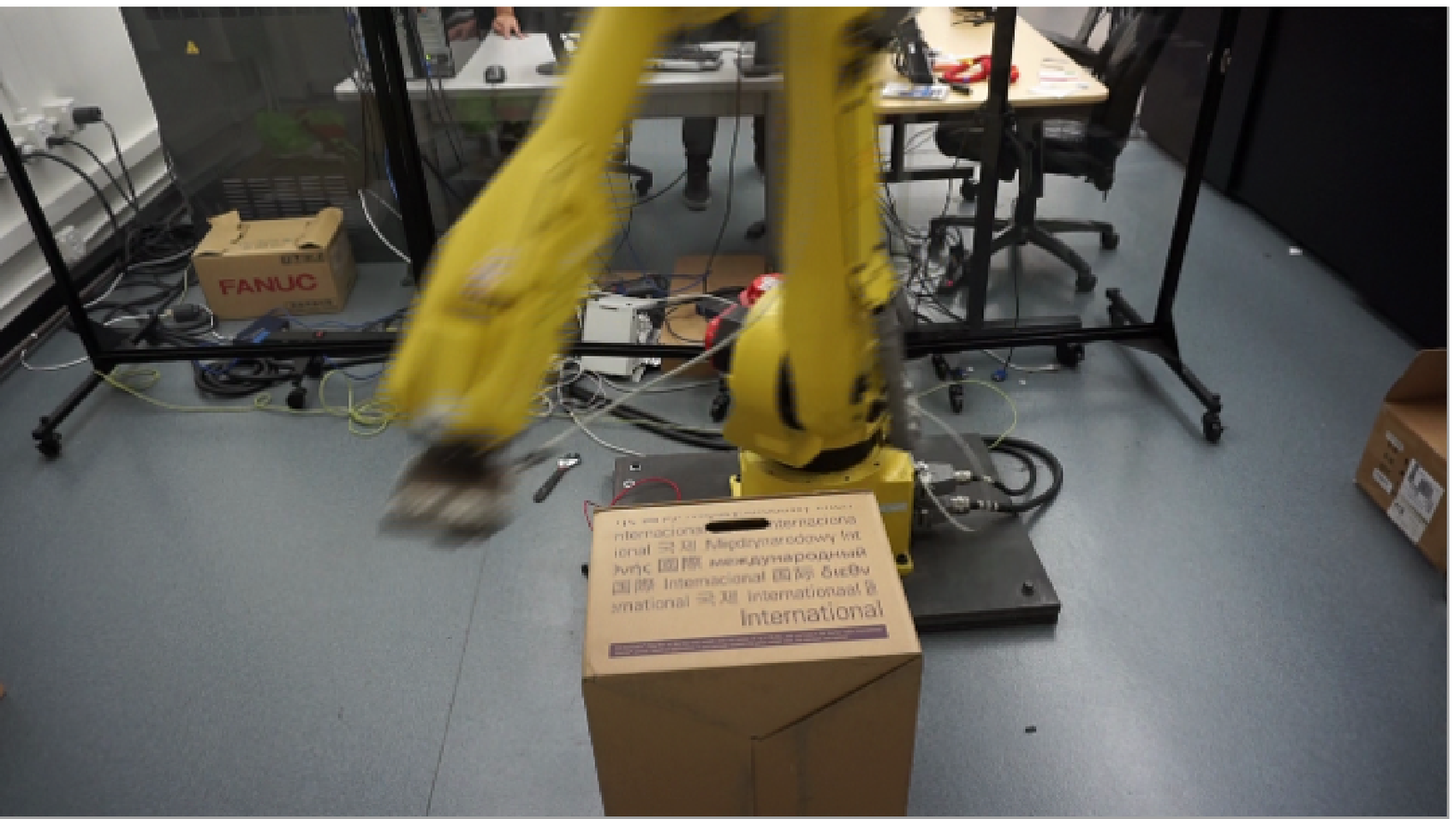}
        \subcaption{t = 0.8 s}
\end{subfigure}
\begin{subfigure}[b]{.22\textwidth}
		\centering
        \includegraphics[width=1\textwidth]{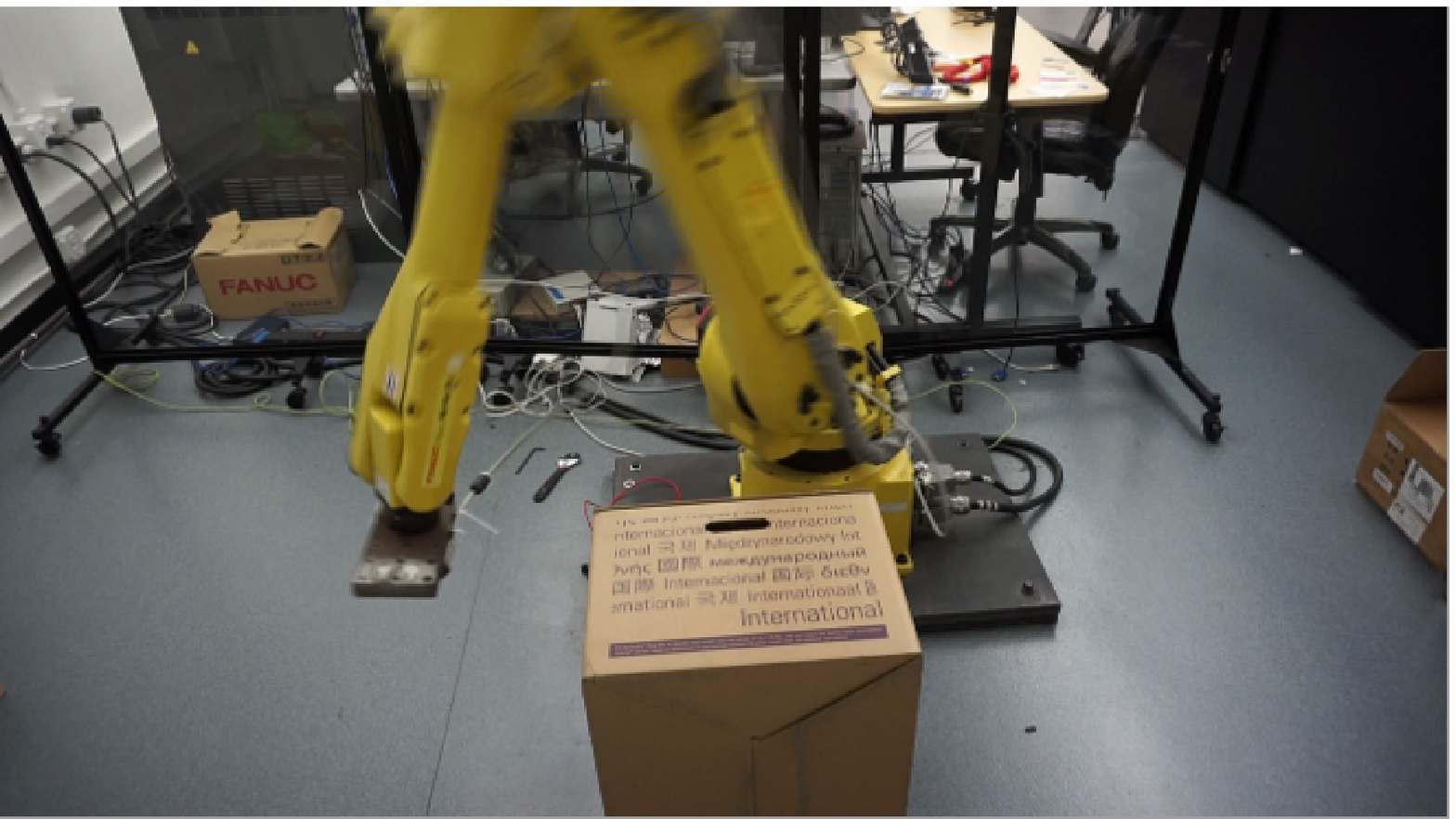}
        \subcaption{t = 0.96 s}
\end{subfigure}
\begin{subfigure}[b]{.22\textwidth}
		\centering
        \includegraphics[width=1\textwidth]{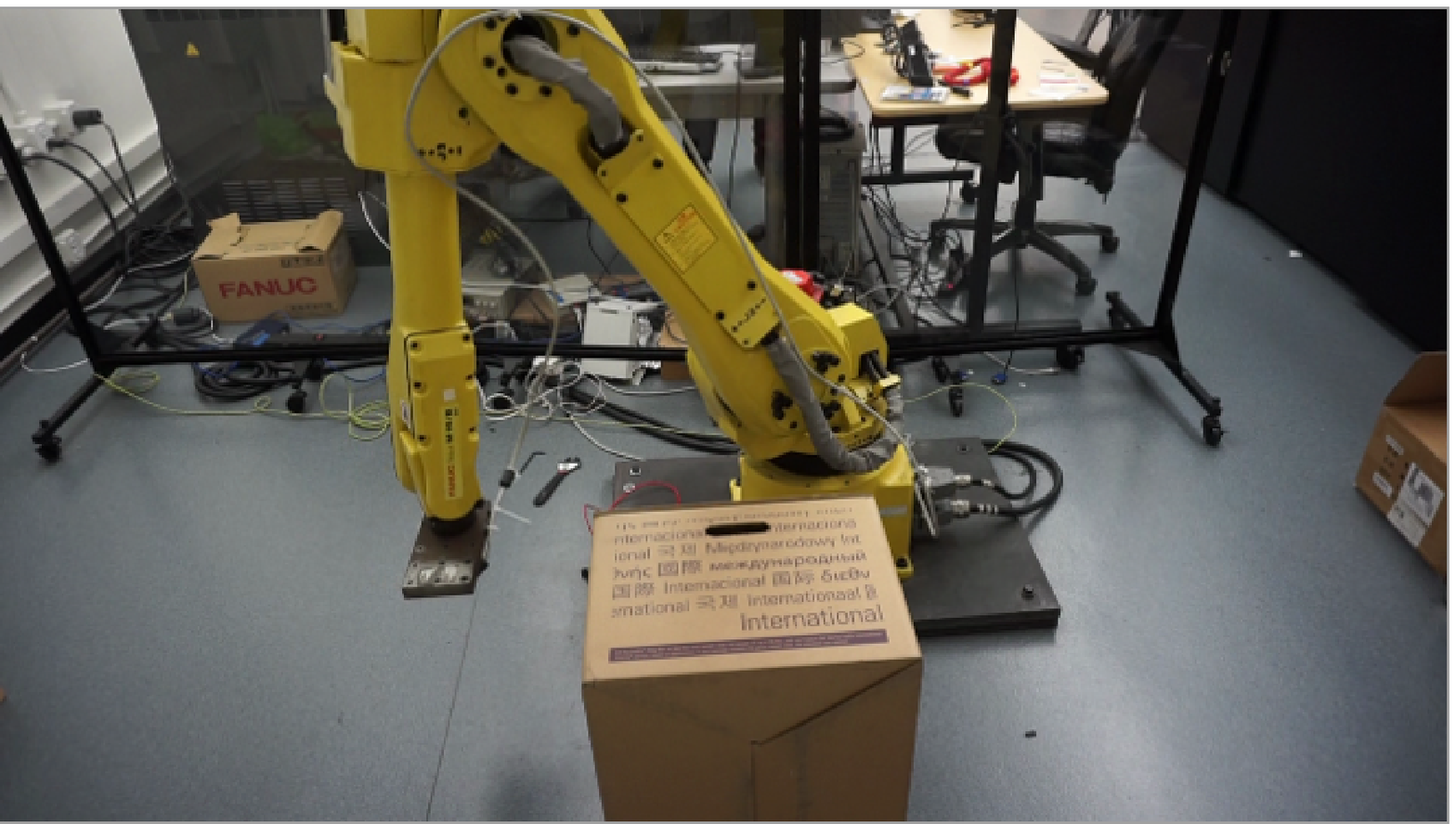}
        \subcaption{t = 1.2 s}
\end{subfigure}
\caption{Actual robot motion in experiment}
\label{expPath}
\end{figure}

The planned optimal trajectory of test case 2 in Fig. \ref{6jnttestInitTarget} has been used as motion reference in experiment at the Mechanical Systems Control laboratory at the University of California, Berkeley. The actual robot motions are captured from video record as shown in Fig. \ref{expPath}. As shown in the figure, the planned motion is collision free. 

\begin{figure}[ht]
\centering
\begin{subfigure}[b]{.45\textwidth}
		\centering
        \includegraphics[width=1\textwidth]{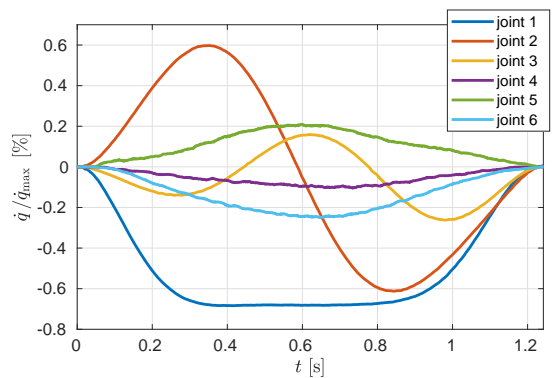}
        \subcaption{Joint velocity}        
\end{subfigure}
\begin{subfigure}[b]{.45\textwidth}
		\centering
        \includegraphics[width=1\textwidth]{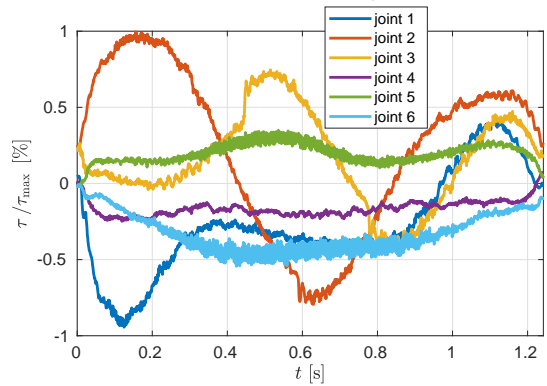}
        \subcaption{Joint torque}
\end{subfigure}
\caption{Measured joint velocity and torque of optimal robot trajectory in experiment}
\label{expVelTrq}
\end{figure}

The scaled joint velocities and joint torques are shown in Fig. \ref{expVelTrq}. As shown in the figure, the planned motion is feasible under conservative actuator limitations. The motion time $t_f$ is adjusted automatically from an initial value 10 s to about 1.2s. It is also observed that the planned motion is close to time-optimal with at least one of the constraints is active.

\section{Conclusion}\label{conclusion}
Motion planning involving complicated robot dynamics and geometric constraints is challenging. Since most approaches decompose motion planning to two subtopics and deal with them separately, only suboptimal solution can be found. This paper presents an optimal control based approach to address the path planning and trajectory planning problems simultaneously. An efficient numerical method for trajectory optimization is proposed as one practical solution for the nonlinear optimal control problem. Numerical results have shown that the motion planning problem can be solved with a short computation time and reasonable accuracy. Experimental results have verified the effectiveness and feasibility of the planning algorithm. It is worth investigating improvements to this approach and exploring possibilities to implement it in different robotic applications.

\bibliographystyle{IEEEtran}
\bibliography{sample}

\end{document}